\documentclass{article}

\usepackage[final]{nips_2018}

\usepackage[utf8]{inputenc} 
\usepackage[T1]{fontenc}    
\usepackage{hyperref}       
\usepackage{url}            
\usepackage{booktabs}       
\usepackage{amsfonts}       
\usepackage{nicefrac}       
\usepackage{microtype}      
\usepackage{graphicx}
\usepackage{xcolor}
\definecolor{orange}{rgb}{0.64,0.16,0.16}
\definecolor{green}{HTML}{2d862d}

\PassOptionsToPackage{numbers}{natbib}
\usepackage{floatrow} 
\newfloatcommand{capbtabbox}{table}[][\FBwidth] 

\title{Natural language understanding for task oriented dialog in the biomedical domain in a low resources context}

\author{
  Antoine Neuraz, Anita Burgun \\
  Department of Biomedical Informatics, H\^{o}pital Necker-Enfants Malades, APHP \\
  INSERM UMRS 1138, Team 22, Paris Descartes, Universit\'{e} Sorbonne Paris Cit\'{e} \\
  \texttt{\{antoine.neuraz,anita.burgun\}@aphp.fr}  \\
    \AND
  Leonardo Campillos Llanos, Sophie Rosset \\
  LIMSI, CNRS, Universit\'{e} Paris Saclay, France \\
 \texttt{\{leonardo.campillos,sophie.rosset\}@limsi.fr}
}

\begin{document}

\maketitle

\begin{abstract}
In the biomedical domain, the lack of sharable datasets often limit the possibility of developing natural language processing systems, especially dialogue applications and natural language understanding models. To overcome this issue, we explore data generation using templates and terminologies and data augmentation approaches. Namely, we report our experiments using paraphrasing and word representations learned on a large EHR corpus with Fasttext and ELMo, to learn a NLU model without any available dataset. We evaluate on a NLU task of natural language queries in EHRs divided in slot-filling and intent classification sub-tasks. On the slot-filling task, we obtain a F-score of 0.76 with the ELMo representation; and on the classification task, a mean F-score of 0.71. Our results show that this method could be used to develop a baseline system. 
\end{abstract}

\section{Introduction}

There is a growing research interest on conversational interfaces for biomedical natural language processing \citep{laranjo2018conversational}. 
Dialogue systems involve several components \citep{jokinen2009spoken}: a natural language understanding (NLU) module, a dialogue manager, a generation module and a module for querying the database. We are interested here in the NLU component, which allows the system to understand user's utterances through the semantic analysis and formalization of queries.

To develop a NLU model using a machine learning approach, one of the first requirements would be a training dataset. This dataset requires user utterances (input of the NLU), along with a formal representation (output of the NLU). The dataset needs to be large enough to be representative of the task. But what if this dataset does not exist? Depending on the task and the language, it is likely that one can not find any suitable training dataset. The biomedical domain is a good use case when it comes to low resources in terms of data, especially in languages other than English. Due to privacy issues, it is difficult to share real world medical data. Dialog systems in the medical domain have been applied for patient counselling on a wide range of topics, from medical conditions to medication intake \citep{azevedo2018}. In most of the systems, interactive capabilities are based on limited, constrained natural language input: for example, users are presented with a menu of multiple-choice questions. In contrast, dialogue systems allow users to access data in a much more natural way---through speech or typed input.\citep{laranjo2018conversational}. 

One solution to overcome the absence of training data is to generate a training dataset based on a few examples and augment it using known terminologies, external knowledge and paraphrases. In this paper, we assess how well models created using such a generated training set perform on real world data and compare their performances on biomedical NLU task in French.

\section{Data generation and augmentation}

In this section, we describe the task and the methods we use to generate the training and development sets. We explain the methods for generating data using templates and terminologies, generating paraphrases of templates using pivot translations and incorporating external knowledge with word embeddings and language models. Figure \ref{fig:expe} details the general schema of this work.

\subsection{Description of the task: NLU in a dialogue task to query EHRs}

The aim of the task is to perform Natural Language Understanding of user input. This step will enable physicians to perform queries in Electronic Health records (EHRs) in natural language. The set of queries that a physician may have about characteristics and results of a patient is broad and diverse, therefore, enabling queries in natural language may help accessing information more efficiently. For this purpose, we asked to medical doctors from a French university hospital some examples of questions they would ask to a dialog system aimed at querying information about biological tests results. We collected a set of 178 questions that we annotated manually as a gold standard.

NLU tasks can usually be divided into 3 sub-tasks: domain classification, intent classification, and slot-filling \citep{tur2011}. For this task in a restricted domain (bio-medicine), we focus on the two latter: slot-filling (sequence labelling) and intent classification. 

\textbf{Sequence labelling}. We distinguish two types of labels: lab mentions (e.g. "créatinine" \textit{creatinin}, "protéine C réactive" \textit{C reactive protein}) and dates (e.g. "27/03/2015", "depuis 3 jours" \textit{since 3 days}). In the training set (generated data, see \ref{section:augmentation}), the number of distinct lab mentions is 336, for a length ranging from 1 to 11 tokens and a median length of 2. There is 28\% of overlapping between the vocabularies of the train set and the test set (real world data). Regarding the date labels, they include actual dates, relative dates and time ranges. The length is more stable with a median of 3, ranging from 0 to 6 tokens. The vocabulary overlap is 38\% between the train and test sets. (see Table \ref{table:mentions})

\textbf{Classification}. There are 4 sub-tasks representing 4 axes of classification. For each utterance, we assign one label per axis. Two axes concern the results of the lab exams (i.e. the type of result (5 categories, e.g. value, evolution, date) and interpretation of the result (5 categories, e.g. normality, value, low, high, presence). The two latter concern temporal aspects (i.e. the time of result (3 categories, e.g. first, last, all) and constraints on time (4 categories, e.g. none, range, date, number).

\subsection{Data generation}
\label{section:augmentation}
Given the lack of suitable dataset for training, we generate a training dataset using a tailored generator. Inspired by \citet{bordes2016}, we developed questions templates: 223 for the core of the question (e.g. "quel est le résultat du dernier <lab mention>", \textit{what is the result of the last <lab mention>}), 23 temporal modifier templates (e.g. "depuis <date|duration|event>" \textit{since <date|duration|event>}) , and a list of 409 mentions of laboratory tests results (hereafter, lab mentions, e.g. "créatinine" \textit{creatinin}, "hémoglobine" \textit{hemoglobin}). Each generated question randomly associates a base template, a temporal modifier template and a lab mention to create unique questions (see Figure \ref{fig:templates}).

\subsection{Data augmentation with paraphrases}
\label{section:para}
Researchers working with data-intensive methods (such as neural networks) already resort to the generation of paraphrases, even for question-answering tasks \citep{D17-1091}.  
We refer the reader to available reviews on methods of paraphrase generation  \citep{androutsopoulos2010survey,madnani2010generating}, including recent advances using neural approaches \citep{iyyer2018adversarial}, and focusing on methods applied for paraphrasing questions. 
A recent approach makes use of the Paraphrase Database (PPDB) \citep{ganitkevitch2013ppdb}, a large multilingual collection of paraphrase pairs (over 100 million pairs for English) with lexical, syntactic and phrasal variations. Another method relies on machine translation \citep{duboue2006answering}. For example, \citep{Zhang:2015:EKC:2887007.2887065} derive paraphrases of key words in questions by translating them to a pivot language (they experimented with 11 languages), then back to the source language.

We use a machine translation method to increase the variability of the training set by producing paraphrases of the question templates. We translate each sentence to one or several languages (pivot languages) and translate the result back to the source language. 
We used the Google Translate API \citep{zotero-5148} for this step. For each template, we randomly select 10 of the 60+ languages available in the API. For each language, we perform the pivot translation and kept the unique paraphrases obtained. We add the paraphrases to the set of templates used for the generation of the datasets.

\subsection{Incorporating latent knowledge}

Using embeddings of words learned on a large domain specific dataset of unlabeled data can be an effective source of latent knowledge~\citep{wang2018}. We use one million of clinical notes from the clinical data warehouse of a local hospital in France. Leveraging this corpus, we compare three types of method: 1) word embeddings (continuous skip-gram) only on the training set (without external knowledge) as a baseline; 2) continuous skip-gram model with sub-word information (i.e. each word is represented as a bag of character n-grams), as implemented in Fasttext \citep{bojanowski2016}, 3) embeddings from language models (ELMo) where the vectors are learned from the internal states of a deep bidirectional language model  as described in \citet{Peters2018}.

\section{Experiments}

We split the question templates and the lab mentions into two sets: training (170 templates and 336 mentions) and  development (53 templates and 73 mentions). From each, we create two datasets by generating paraphrases (see section \ref{section:para}). We generate 16,000 utterances for the training set (80\%) and 4,000 for the development set (20\%) using templates without paraphrases, and the same quantities for the sets with paraphrases (Table S2). The test set (real world data) is kept aside for the evaluations. 

A usual way of producing specialized NLU systems is to elaborate rule-based algorithms to perform the semantic parsing of user's utterances \citep{weston2015}. However, developing such system can be time consuming and is often difficult to maintain. Most of modern NLU systems use statistical learning models to perform this task \citep{young2013a}. Before the raising of neural based systems, state of the art systems used conditional random fields (CRF) \citep{lafferty2001conditional}. Nowadays, these systems tend to be outperformed by neural based approaches, such as convolutional neural networks (CNN) and recurrent neural networks (RNN). On the task of sequence labelling, RNNs and more specifically long short term memory units (LSTM) \citep{hochreiter1997long} are the most used. More recent work combine bidirectional LSTMs (biLSTM) and CRF \citep{lample2016neural}.

To assess the capacity of the models to generalize to new data, we evaluate three types of models for this task: CRF, bidirectional LSTM (biLSTM), and a combination of biLSTM and CRF \citep{lample2016neural}. The input layer is fed with the generated questions from templates only or from tem- plates with paraphrases. The embeddings are learned either directly on the training set (no external knowledge), or on clinical notes using Fasttext or ELMo. For each combination, we test three different models: CRF, biLSTM and biLSTM+CRF. The details of the models and the tuning parameters are described in the supplementary materials (section \ref{tuning}). All the results are reported in terms of weighted F-measure, computed using 10 repetitions of five fold cross-validation over the test set. 

\section{Results and discussion}

Overall, the best results on sequence labelling and classification tasks are obtained with the models including ELMo representations as the embeddings used to inject external knowledge. On the sequence labelling task, models with ELMo-biLSTM and ELMo-biLSTM-CRF obtained a F1-score of 0.76(95\%CI [0.74-0.77]) and 0.77 (95\%CI [0.76-0.79]) respectively (see Table \ref{table:labseq}, Figure \ref{fig:vasequence}). On the classification task, the best results are obtained with ELMo on three of the four sub-tasks and with Fasttext-paraphrases on the forth one (see Table \ref{table:classif}, Figure \ref{fig:vaclassif}). 

On the sequence labelling task, adding latent knowledge with FastText or ELMo using a million clinical records increases the generalizability of the models regardless of the type of the downstream model. Models with ELMo have an average F1-score of $0.75 (\pm 0.05)$ , with FastText $0.66 (\pm 0.06)$ and without external knowledge $0.55 (\pm 0.12)$. Adding paraphrases to the templates does not improve the results on this task and even tends to lower the results: ELMo $0.76 (\pm 0.05)$ without paraphrases and $0.74 (\pm 0.05)$ with; FastText $0.66 (\pm 0.06)$ versus $0.66 (\pm 0.06)$; no external embedding $0.55 (\pm 0.10)$ versus $0.54 (\pm 0.14)$. Regarding the type of model, biLSTM and biLSTM-CRF perform better than CRF only with F1-scores of $0.69 (\pm 0.08)$, $0.68 (\pm 0.08)$ and $0.59 (\pm 0.15)$, respectively. 

On the classification tasks, we also observe better results with ELMo and fastText than without external embedding: mean F1-scores of $0.68 (\pm 0.12)$ with ELmo, $0.66 (\pm 0.11)$ with FastText and $0.61 (\pm 0.12)$ without external embedding. Unlike for the sequence labelling task, adding paraphrases to the training set tends to give better results with F1-scores of $0.63 (\pm 0.13)$ without and $0.67 (\pm 0.10)$.  

Interestingly, the results obtained with the best models on each task show that it is possible to use our method to provide a baseline system for NLU tasks in the absence of a pre-existing data. Our results not only confirm those by \citet{wang2018} regarding the interest of incorporating external knowledge using a large domain specific corpus; our  outcomes also highlight the interest of using language models instead of only embeddings to incorporate this knowledge. In our study, the results using ELMo are systematically better than those with FastText although the models were learned on the same data. This may come from the better representation of the context in ELMo compared to FastText. FastText takes into account the tokens in the specified window, which can be described as a "bag of context". But ELMo is a language model and considers the full context of a token (at the sentence level). Of note, this sequence labelling task is not very complex, given the number of different labels. The results on a task with more labels might be lower. Moreover, the results with the paraphrases are more difficult to interpret: they are slightly better on the classification tasks but not on the sequence labelling task.
This might come from the method of pivot translation used for producing this paraphrases. Indeed, the quality of the produced paraphrases may not be sufficient for the task. Using  more sophisticated methods of paraphrasing could lead to different results.

\section{Conclusion}

NLU models learned on the data generated using the proposed method achieve interesting performances. These methods can be considered to learn a baseline model allowing to bootstrap a dialog system and start collecting data from end users. We are interested in exploring to which extent other sources to train embeddings (e.g. medical, non-clinical texts) yield similar results. It would also be interesting to conduct similar experiments in related tasks where data are scarce (e.g. NLU in dialogue systems for patient counselling or virtual patients).

\begin{table}[ht]
\setlength{\tabcolsep}{1pt} 
\begin{floatrow} 
\capbtabbox[.51\textwidth]{%
\begin{tabular}{lc}
    \toprule
    \textbf{Model}                              & \textbf{F1-score [95\%CI]}         \\
    \midrule
    CRF + para                             & .37 [.35-.39]             \\
    CRF                                    & .43 [.42-.44]             \\
    BiLSTM                                 & .59 [.58-.61]            \\
    BiLSTM + para                          & .66 [.65-.68]             \\
    BiLSTM + CRF                           & .62 [.60-.63]             \\
    BiLSTM + CRF + para                    & .60 [.59-.62]             \\
    CRF + FastText + para                  & .62 [.60-.63]             \\
    CRF + FastText                         & .62 [.61-.64]             \\
    BiLSTM + CRF + FastText                & .67 [.65-.68]             \\
    BiLSTM + CRF + FastText + para         & .67 [.65-.69]             \\
    BiLSTM + FastText                      & .69 [.67-.71]             \\
    BiLSTM + FastText + para               & .68 [.67-.70]             \\
    BiLSTM + CRF + ELMO + para             & .73 [.71-.74]              \\
    BiLSTM + ELMO + para                   & .74 [.72-.75]             \\
    CRF + ELMO                             & .75 [.73-.76]             \\
    CRF + ELMO + para                      & .75 [.74-.76]             \\
    BiLSTM + CRF + ELMO                    & \textbf{.76 [.74-.77]}    \\
    BiLSTM + ELMO                          & \textbf{.77 [.76-.79]}    \\
    \bottomrule
\end{tabular}
}{%
\caption{Results of the experiments for the sequence labelling task. para = paraphrases} 
\label{table:labseq} 
} 
\capbtabbox[.45\textwidth]{%
\begin{tabular}{lcc}
    \toprule
    \textbf{Model}                       & \multicolumn{2}{c}{\textbf{Sub-task (F1-score) }} \\
                                 & \small{Type}           & \small{Interpretation}   \\ \midrule
    train-set embedding          &.64 [.62-.67]          & .65 [.63-.68]                    \\
    EHR ELMO                     & \textbf{.70 [.68-.73]} & .64 [.62-.67]                 \\
    EHR FastText                 & \.69 [.67-.72]  & .68 [.66-.70]          \\
    train-set + para             & .62 [.59-.64]      & \textbf{.71 [.69-.72]}                 \\
    EHR FastText + para          & .64 [.62-.66]      & .68 [.66-.70]    \\
    EHR ELMO + para              & .65 [.62-.67]      & .68 [.65-.70]     \\ 
    \bottomrule
    \toprule
    \textbf{Model}                       & \multicolumn{2}{c}{\textbf{Sub-task (F1-score) }} \\
                                 & \small{Time}           & \small{Time constraint}   \\ \midrule
    train-set embedding          &.68 [.65-.70]           &.40 [.38-.42]             \\
    EHR ELMO                     & \textbf{.77 [.75-.79]} & .41 [.40-.44]         \\
    EHR FastText                 & .70 [.68-.72]          & .42 [.40-.44]        \\
    train-set + para             & .72 [.70-.74]          & .53 [.50-.55]             \\
    EHR FastText + para          & .74 [.72-.76]          & \textbf{.72 [.69-.74]}     \\
    EHR ELMO + para              & .75 [.73-.77] & .72 [.70-.74]     \\ 
    \bottomrule
\end{tabular}
}{%
\caption{Results of the experiments for the classification task. para = paraphrases} 
\label{table:classif} 
} 
\end{floatrow} 
\end{table}
  

\clearpage
\medskip

\small

\bibliographystyle{abbrvnat}
\bibliography{biblio}

\clearpage
\section*{\large Supplementary material}

\setcounter{figure}{0}
\setcounter{table}{0}
\setcounter{section}{0}
\renewcommand{\thefigure}{S\arabic{figure}}
\renewcommand{\thetable}{S\arabic{table}}
\renewcommand{\thesection}{S\arabic{section}}

\begin{table}[ht]
\small
\centering
\caption{Description of the term mentions}
\begin{tabular}{lcc}
\toprule
       & date          & lab mention   \\
\midrule
mentions in the test set      & 34            & 177           \\
Median length[min-max] & 3 {[}0 - 6{]} & 2 {[}1 - 11{]} \\
vocabulary in train set  & 1,364 &   451 \\
vocabulary in test set(intersection with train) & 58(0.38) & 250(0.28) \\
\bottomrule
\end{tabular}
\label{table:mentions}
\end{table}

\begin{table}[ht]
\begin{center}
\caption{Characteristics of the datasets. (*) with paraphrases. OOV = out of vocabulary}
\footnotesize\setlength{\tabcolsep}{2.5pt}
\begin{tabular}{lcccccc}
      \toprule
           	& Utterances & Templates & Lab mentions & Words (*)  	& OOVs (*) & Perplexity (*)  \\
      \midrule
training   	& 16,000  	 & 170 		 & 336 & 144,850 (140,492)  & -    & -  \\
development & 4,000  	 & 53 		 & 73 	 & 36,211 (36,211) 	& 4,724 (4,544)  & 137.5 (171.1)  \\
test 		& 178  		 & - 		 & -  & 1,579 	& 467 (390) & 194.5 (240.1) \\
      \bottomrule

\end{tabular}
\end{center}
\label{table:dt}
\end{table}

\begin{figure}[ht]
\begin{center}
\includegraphics{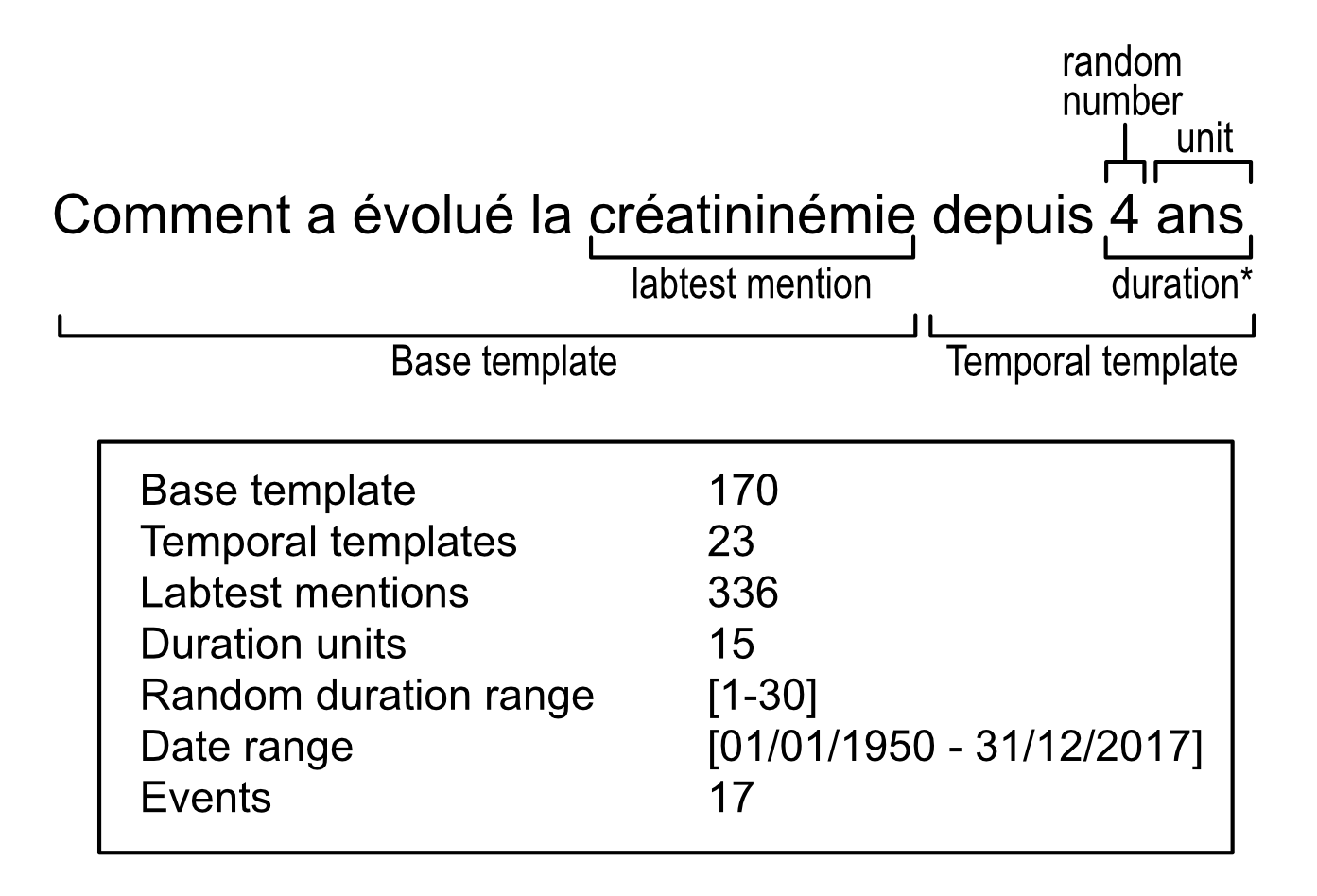} 
\caption{Combinations of templates and modifiers.}
\raggedright{* duration can be replaced by date or event. The example means: `How has creatininemia evolved in the last 4 years'}
\label{fig:templates}
\end{center}
\end{figure}

\begin{figure}[ht]
\begin{center}
\includegraphics[scale=0.7]{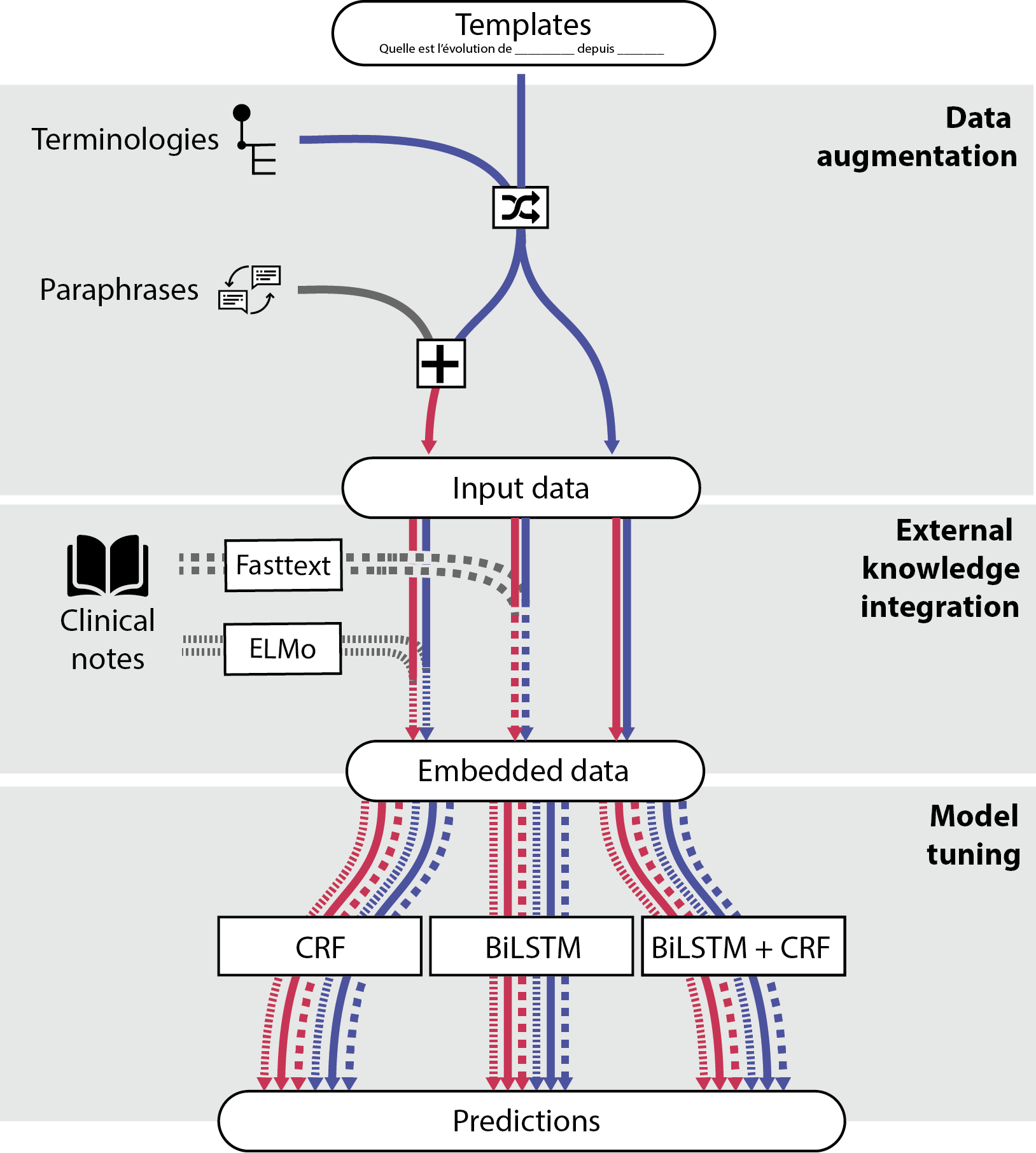}
\caption{Experiences flow.}
\label{fig:expe}
\end{center}
\end{figure}

\section{Tuning parameters}
\label{tuning}
For each model (except ELMo) we added some standard features to the input: normalized lemmas and part-of-speech (POS) tagging.
Then, the sequence labelling part of the model was constituted of a CRF only or 2 layers of biLSTM or 2 layers of biLSTM followed by a CRF. The tuning parameters were: the dimension of the embeddings (50, 100, 300) except for ELMo (fixed to the default dimension), the number of units units in the biLSTM (64, 128, 256), the fraction of dropout after the embedding layer and after the LSTM layers (0.1, 0.2, 0.3, 0.4, 0.5). Regarding the classification part of the model, it constituted of a 1 dimensional convolution layer (2 to 5 filter kernel size and 50 to 250 filters, ReLu activation) followed by a max-pooling layer. Models were tuned using a random sample of parameters. The optimization function was ADAM. All the models were implemented using Keras \citep{chollet2015keras} with a Tensorflow \citep{tensorflow2015-whitepaper} backend.

  \begin{figure}[ht]
    \begin{center}
    \includegraphics[scale=0.25]{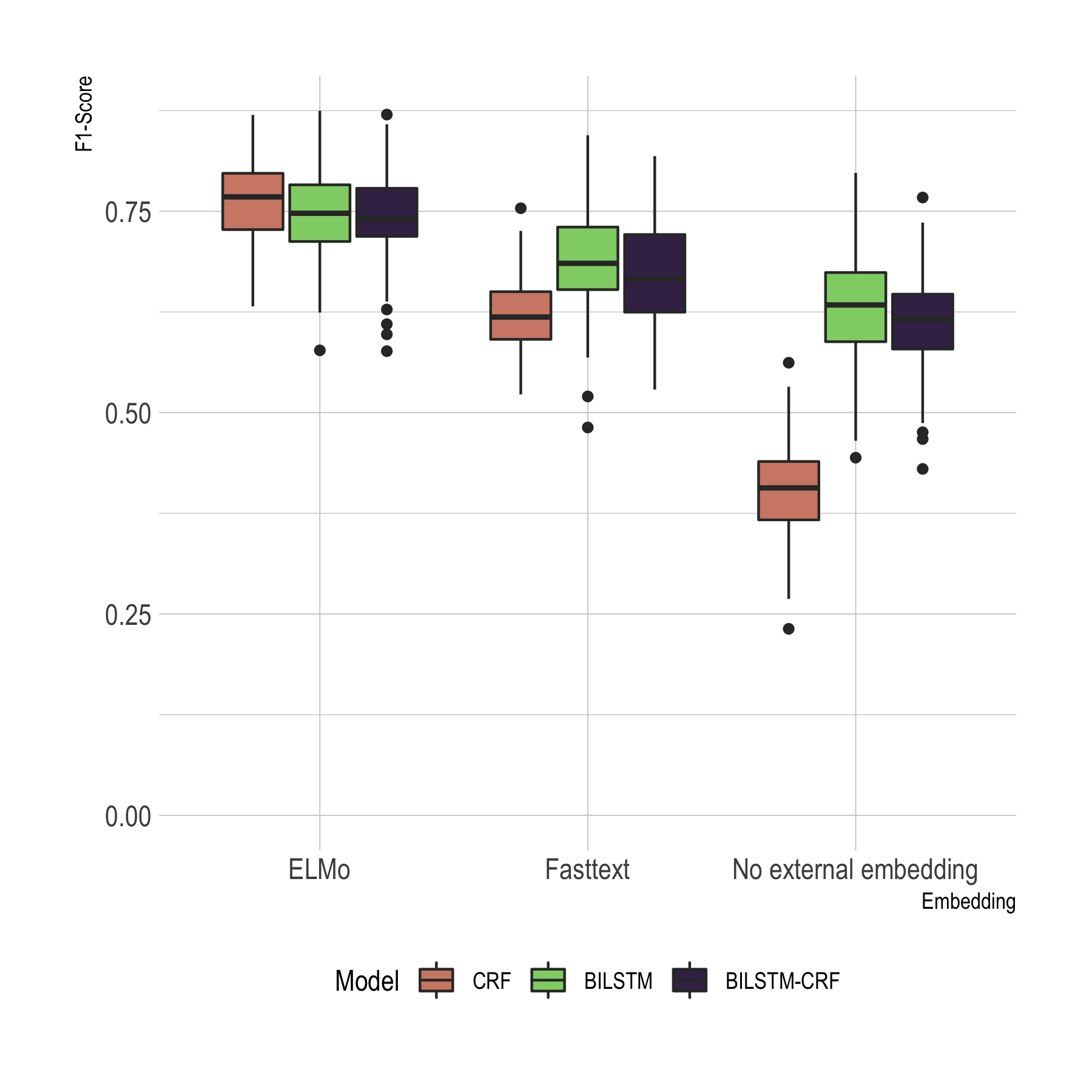}
    \caption{VA Task - sequence labelling}
    \label{fig:vasequence}
    \end{center}
    \end{figure}
      
    \begin{figure}[h]
    \begin{center}
    \includegraphics[scale=0.25]{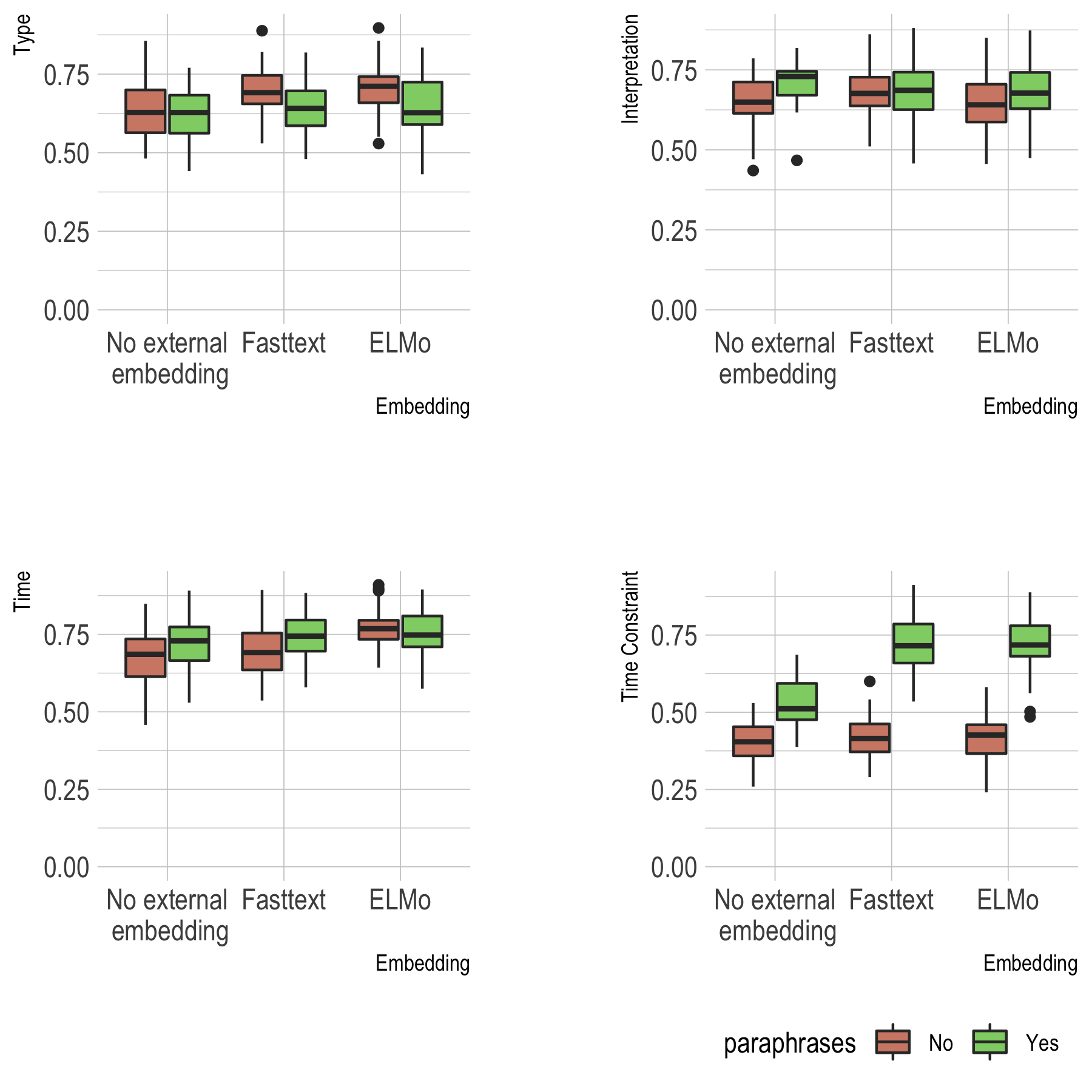}
    \caption{VA Task - classification}
    \label{fig:vaclassif}
    \end{center}
    \end{figure}

\end{document}